\documentclass{ieeeaccess}
\usepackage{csquotes}
\usepackage{cite}
\usepackage{amsmath,amssymb,amsfonts}
\usepackage{algorithmic}
\usepackage{graphicx}
\usepackage{textcomp}
\usepackage{subfig}
\usepackage{bbold}
\def\BibTeX{{\rm B\kern-.05em{\sc i\kern-.025em b}\kern-.08em
    T\kern-.1667em\lower.7ex\hbox{E}\kern-.125emX}}
\begin{document}
\history{Date of publication xxxx 00, 0000, date of current version xxxx 00, 0000.}
\doi{10.1109/ACCESS.2017.DOI}

\title{Improving Health Mentioning Classification of Tweets using Contrastive Adversarial Training}
\author{\uppercase{Pervaiz Iqbal Khan}\authorrefmark{1,2}, 
\uppercase{Shoaib Ahmed Siddiqui \authorrefmark{1,2}},
\uppercase{Imran Razzak \authorrefmark{3}},
\uppercase{Andreas Dengel \authorrefmark{1,2}, and }
\uppercase{Sheraz Ahmed \authorrefmark{1} }
}
\address[1]{German Research Center for Artificial Intelligence (DFKI), 67663 Kaiserslautern, Germany}
\address[2]{TU Kaiserslautern, 67663 Kaiserslautern, Germany}
\address[3]{UNSW, Sydney, Austrailia}
\tfootnote{}

\markboth
{Pervaiz \headeretal:Improving Health Mentioning Classification of Tweets using Contrastive Adversarial Training}
{Pervaiz \headeretal: Improving Health Mentioning Classification of Tweets using Contrastive Adversarial Training}

\corresp{Corresponding author: Pervaiz Iqbal Khan (pervaiz.khan@dfki.de).}

\begin{abstract}
Health mentioning classification (HMC) classifies an input text as health mention or not. Figurative and non-health mention of disease words makes the classification task challenging. Learning the context of the input text is the key to this problem. The idea is to learn word representation by its surrounding words and utilize emojis in the text to help improve the classification results. In this paper, we improve the word representation of the input text using adversarial training that acts as a regularizer during fine-tuning of the model. We generate adversarial examples by perturbing the embeddings of the model and then train the model on a pair of clean and adversarial examples. Additionally, we utilize contrastive loss that pushes a pair of clean and perturbed examples close to each other and other examples away in the representation space. We train and evaluate the method on an extended version of the publicly available PHM2017 dataset. Experiments show an improvement of $1.0\%$ over BERT\textsubscript{Large} baseline and $0.6\%$ over RoBERTa\textsubscript{Large} baseline, whereas $5.8\%$ over the state-of-the-art in terms of F1 score. Furthermore, we provide a brief analysis of the results by utilizing the power of explainable AI.
\end{abstract}


\titlepgskip=-15pt

\maketitle

\section{Introduction}
\label{sec:introduction}
\PARstart{H}{ealth} mentioning classification (HMC) deals with the classification of a given piece of a text as health mention or not. This helps in the early detection and tracking of a pandemic that enables health departments and authorities in managing the resources and controlling the situation. The input text is gathered from the social media platforms such as Twitter, Facebook, Reddit, etc. The collection process involves crawling the aforementioned platforms based on keywords containing disease names. The keyword-based data collection does not consider the context of the text and hence contains irrelevant data. For example a tweet \enquote{I made such a great bowl of soup I think I cured my own depression} contains a disease of \enquote{depression} but this is used figuratively. Another tweet \enquote{Hearing people cough makes me angry. I cannot explain it} contains \enquote{cough} in it but this does not show that a person is having a cough. Non-health and a figurative mention of disease words in these cases pose challenges to the HMC. So, the question arises how to address these challenges? One way is to consider surrounding words of the disease words that will give the context of the text. Another way is to leverage the emojis in the text as figurative mentioning text may contain smileys whereas the actual disease mentioning text may contain emojis of sad faces, etc.

Transformed methods \cite{vaswani2017attention} are good at capturing the contextual meanings of the words and have shown success in many natural language processing (NLP) tasks. BERT \cite{devlin2018bert} is a transformer model pre-trained on a large unlabelled text corpus for language understanding, and can be fine-tuned on downstream tasks such as text classification\cite{sun2019fine}. It considers the words on the left and right sides of a given word while learning a representation for it. In this way, it achieves the contextual representation of a given word. BERT randomly masks 15\% of the tokens in the corpus and then tries to predict masked tokens during the training process. RoBERTa \cite{liu2019roberta} is an improvement over the BERT using dynamic masking of words instead of static 15\% masking of the words. Further, it is trained on 1000\% more data than BERT. Existing health mentioning classification tasks use both non-contextual, and contextual representations for the given text\cite{biddle2020leveraging, jiang2018identifying,karisani2018did,iyer2019figurative,khan2020improving}. However, contextual representations have improved the performance of the classifier over non-contextual representations. Some methods use emojis present in the tweet text for the classification task. \cite{biddle2020leveraging} extracts the sentiment information from the given tweet and passes it as an additional feature with textual features.\cite{khan2020improving} converts emojis into text using Python library and then utilizes this emoji text as a part of tweet text.

Adversarial training (AT) \cite{goodfellow2014explaining} works as a regularizer and improves the robustness of the model against adversarial examples. The key idea is to add a gradient-based perturbation to the input examples, and then train the model on both clean and perturbed examples. In contrast to images, this technique is not directly applicable to text data.\cite{miyato2016adversarial} applies perturbations to word embeddings for the task of text classification. \cite{chen2020simple} utilizes a contrastive loss for learning features in computer vision (CV). The idea is, the input image is perturbed by adding some augmentation, and during training contrastive loss pushes both clean and augmented examples together while it pushes other examples away from these examples. Contrastive loss helps model learning noise invariant image feature representation.\cite{pan2021improved} proposes contrastive adversarial for text classification that improves the performance over the baseline methods. In this work, we propose contrastive adversarial training on the task of HMC, additionally using contrastive loss during the fine-tuning of models. Specifically, we add perturbation to the embedding matrix of BERT and RoBERTa using Fast Gradient Sign Method (FGSM)\cite{goodfellow2014explaining}. Then we train both the clean and perturbed training examples simultaneously. Our method outperforms both BERT\textsubscript{Large} and RoBERTa\textsubscript{Large} baseline methods as well as achieves state-of-the-art (SOTA) performance on HMC task. Generally, deep learning models are black regarded as black boxes, i.e., it is not clear what information in the input influences the models to make their decisions. European Union adopted new regulations to implement a \enquote{right to explanation} which means a user can ask for the explanation of a decision made by the algorithm\cite{goodman2017european}. Explainable AI focuses on explaining the decisions made by algorithms. In this paper, we leverage explainable AI capabilities to visualize the words that contribute to the model decision. The main contributions of this paper are:
\begin{itemize}
    \item We show that using contrastive adversarial training as a regularizer improves the performance of the model.
    \item We significantly improve HMC performance over SOTA.
    \item We provide the analysis of improvement in the performance by leveraging the power of explainable AI. 
\end{itemize}

The rest of the paper is organized as follows: In section \ref{ls}, we discuss the related work, whereas in section \ref{method} we present our method for HMC. In section \ref{exp}, we give experimentation detail. In section \ref{raa}, we present results and analysis of the experiments. In section \ref{conc}, we provide the conclusion of the paper.

\section{Related Work}\label{ls}
In this section, we discuss existing work in the literature related to adversarial training, contrastive learning, and health mention classification of tweets.

\subsection{Adversarial Training}
Adversarial Training (AT) has been studied in many supervised classification tasks such as object detection\cite{chen2018robust,song2018physical,xie2017adversarial}, object segmentation\cite{xie2017adversarial,arnab2018robustness} and image classification\cite{goodfellow2014explaining,papernot2016limitations,su2019one}. AT is the process of training the model to defend against malicious \enquote{attacks} and increase network robustness. AT involves the training of the model simultaneously with adversarial and clean examples. These malicious attacks are generated by perturbing the original input examples, so that the model predicts the wrong class label \cite{madry2017towards, chakraborty2018adversarial} for them. Fast Gradient Sign Method (FGSM) proposed in\cite{goodfellow2014explaining} is the method for generating adversarial examples for images. \cite{miyato2016adversarial} extends FGSM to NLP tasks such that it perturbs word embeddings instead of original text inputs and applies the method to both supervised and semi-supervised settings with Virtual Adversarial Training (VAT)\cite{miyato2015distributional} for the latter. Recent works propose to add perturbations to the attention mechanism of transformer-based methods\cite{kitada2021attention,kitada2021making,zhu2019freelb}. Compared to single-step FGSM, \cite{madry2017towards} applies the multi-step approach to generate adversarial examples that proves more effective as compared to single-step FGSM, however it increases the computational cost due to the inner loop that iteratively calculates the perturbations. \cite{shafahi2019adversarial} proposes free adversarial training, where the inner loop calculates the perturbation as well as gradients with respect to the model parameters and updates the model parameters. \cite{zhu2019freelb} also uses the free AT algorithm and adds gradient accumulation to achieve a larger effective batch. It also applies perturbations to word embeddings of LSTM and BERT-based models similar to \cite{miyato2016adversarial}. In our work, we generate adversarial examples using one-step FGSM and perform contrastive learning with clean examples to learn the representations for the input examples.

\subsection{Contrastive Learning}
Self-supervised contrastive learning methods, such as MoCo\cite{he2020momentum} and SimCLR\cite{chen2020simple} have narrowed down the performance gap between self-supervised learning and fully-supervised methods on the ImageNet\cite{deng2009imagenet} dataset. It has also been applied successfully in the natural language processing (NLP) domain. The main idea in contrastive learning is to create positive pair to train the models. Various methods have been used to create these pairs. \cite{fang2020cert} uses back-translation to generate another view of the input data.\cite{wu2020clear} uses the word and span deletion, reordering, and substitution of words, whereas \cite{meng2021coco} crops and masks sequences from an auxiliary Transformer to create positive pairs. \cite{gunel2020supervised} performs supervised contrastive learning\cite{khosla2020supervised} by treating training examples of the same class as positive pairs. To generate positive examples, \cite{gao2021simcse} uses different dropout masks on the same data and treats premises and their corresponding hypotheses as positive pairs and contradictions as hard negatives in the NLI datasets\cite{bowman2015large, williams2017broad}. In our work, we treat an original input and its adversarial example as a positive pair and all other examples in the training batch as negative examples. We further use contrastive loss during fine-tuning of models that pushes the positive pairs close to each other and negative examples away from them in the representation space.

\subsection{Health Mention Classification}
\cite{karisani2018did} presents a new method namely Word Embedding Space Partitioning and Distortion (WESPAD) for health mention classification on Twitter data. WESPAD first learns to partition and then distort word representations, which acts as a regularizer and adds generalization capabilities to the model. This method also solves the problem of little training examples for the positive health mentions in the dataset. Although, this method improves the classification accuracy, distorting the original word embedding causes information loss. \cite{jiang2018identifying} uses non-contextual word embeddings for tweet health classification. It applies the preprocessing on the given tweet and extracts non-contextual word representations from it, and then passes these representations to Long Short-Term Memory Networks (LSTMs) \cite{hochreiter1997long}. LSTMs-based classifier outperforms Support Vector Machines (SVM), K-Nearest Neighbor (KNN), and Decision Trees. \cite{iyer2019figurative} uses a two-stepped approach for the tweet classification. First, it detects whether the disease word is mentioned figuratively or not, and then, it uses this information as a new binary feature combined with other features and applies a convolutional neural network (CNN) for the classification. Usage of this additional feature improves the classification results. This method does not work well on figurative mention tweets, especially the disease word \enquote{heart attack}, one of the most widely used words in the figurative sense. \cite{biddle2020leveraging} adds 14k new tweets to the existing health-mention dataset \enquote{PHM2017}\cite{karisani2018did
}. It also uses emojis by converting them into string representations using the Python library. As a pre-processing, it normalizes the URL and user mentions in the tweet. This work experiments with both non-contextual representations such as word2vec \cite{mikolov2013distributed} as well as with contextual representations like ELMO \cite{peters2018deep} and BERT \cite{devlin2018bert} and incorporates sentiment information using WordNet \cite{baccianella2010sentiwordnet}, VAD \cite{mohammad2018obtaining}, and ULMFit \cite{howard2018universal}. It combines the output of the Bi-LSTM \cite{graves2013speech} with sentiment information to produce a final binary output that represents classification results. Experiments show that combining BERT and VAD outperforms other methods. \cite{khan2020improving} uses permutation-based word representation method \cite{yang2019xlnet} for health mention classification and leverages the emojis as a part of the tweet text by converting them into a text representation. 

In this paper, we exploit the adversarial training combined with the contrastive learning on the task of HMC. Results show that adversarial training combined with contrastive learning consistently improves the classification score.
\section{Method}\label{method}
\begin{figure*}[!t] \centering{\includegraphics[width=12cm ,keepaspectratio]{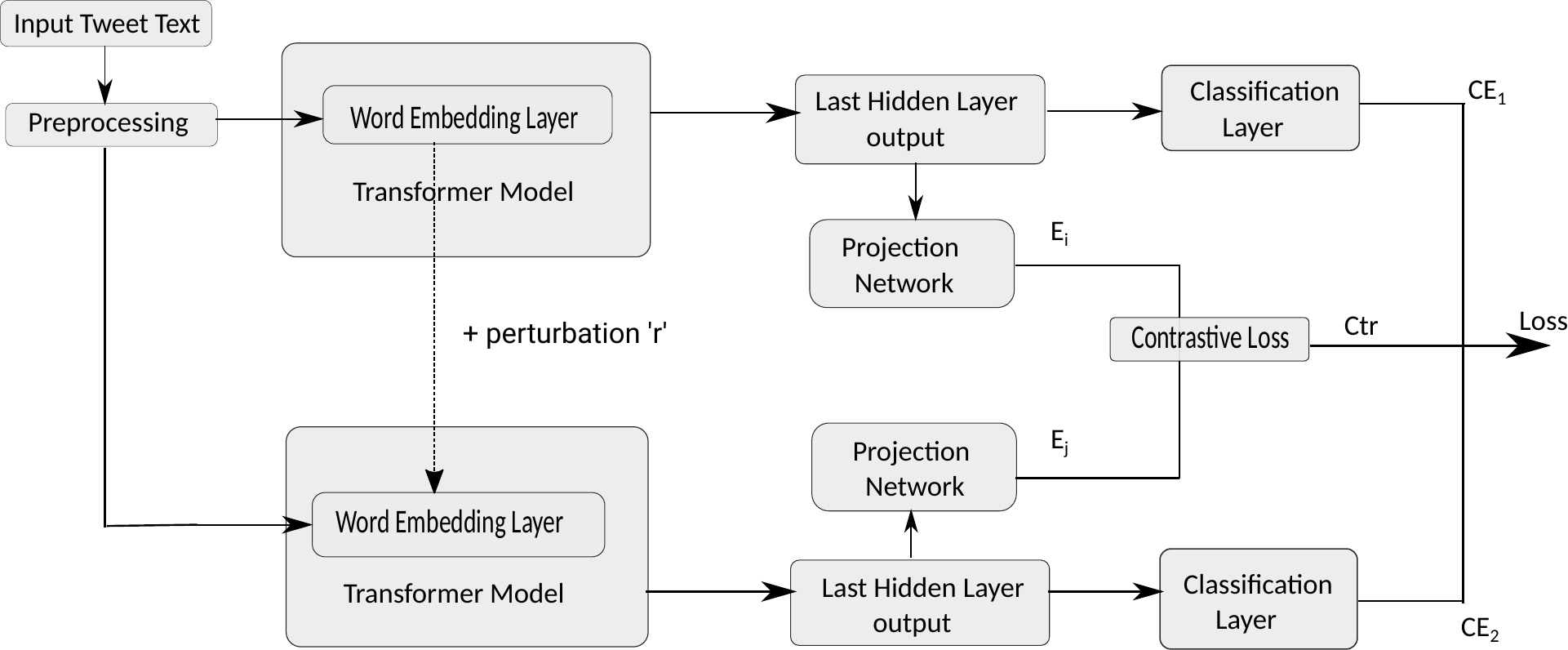}} \caption{Method for contrastive adversarial training. Every input example goes through pre-processing step, then it completes its forward pass through transformer model. We perturb the embedding matrix using Fast Gradient Sign Method (FGSM) to generate adversarial examples. Then we train both clean and perturbed examples. We also utilize contrastive loss represented as `Ctr' that pushes clean and its adversarial example close to each other in the representation space, and it pushes negative examples apart. The final loss is the weighted sum of two cross-entropy losses and one contrastive loss.} \label{method-diagram} \end{figure*}

In this section, we describe the basics of the transformer-based encoder for text classification. Then we discuss adversarial training and contrastive loss. Finally, we discuss how to combine these ideas to improve health mentioning classification score. Figure \ref{method-diagram} shows the overall architecture of the model. 
\subsection{Transformers Basics}
Let $\{x_i,y_i\}_{i=1,...,N}$ be training examples in the dataset and `M' be a pre-trained model such as BERT or RoBERTa. Each training example is represented as tokens of sequences i.e. $x_i=[CLS, t_1,t_2,...., t_T,SEP]$ as input to M that outputs contextual token representations $[h^L_{CLS}, h^L_{1},h^L_{2},...., h^L_{T},h^L_{SEP}],$ where `L' denotes number of layers in `M'. \\

To fine-tune pre-trained model `M', a softmax classifer is added as a final layer that takes the hidden representation $h_{CLS}$ of the $[CLS]$ token. A model `M' is trained by minimizing cross entropy loss:\\
 \begin{equation}\label{eq:loss}
\mathcal{L}_{CE} = - {\dfrac{1}{N}}\sum_{i=1}^{N}\sum_{c=1}^{C}y_{i,c}log(p(y_i,c|h^{i}_{[CLS]}))
    \end{equation}

where  `C' denotes the number of classes in the dataset, and `N' is the number of training examples in a batch. 

\subsection{Adversarial Training}
Adversarial training involves perturbing the inputs to the model that cause misclassifications. Fast Gradient Sign Method (FGSM) is proposed by \cite{goodfellow2014explaining} to generate perturbed examples. The model is trained on both clean and adversarial examples that improve the model's robustness against adversarial attacks. Let, `r' be the small perturbation to the input example $x_i$, and $y_i$ be the ground truth.  Then we maximize the loss function:

\begin{equation}\label{eq:1}
    max\mathcal{L}(f_{\theta}(x_i+r), y_i), s.t.\| r \|_{\infty} < \epsilon \text{,    where    } \epsilon > 0
\end{equation}

where $\mathcal{L}(f_{\theta}(x_i+r), y_i)$ is the loss function and $f_\theta$ is the neural network parameterized by $\theta$.
\\

To produce the perturbation `r', Equation (\ref{eq:1}) can be simplified as follows:

\begin{equation}\label{eq:2}
 r = -\epsilon sign(\nabla_{x_i}\mathcal{L}(f_\theta (x_i), y_i))
 \end{equation}

To generate adversarial examples, similar to\cite{miyato2016adversarial} we perturb the embedding matrix $E \in \mathbb{R}^{d_v \times d_h}$ where $d_h$ is hidden unit size and $d_v$ is vocabulary size in the transformer model `M'. At the end of each forward pass, we calculate the gradient of the loss function given in equation (\ref{eq:loss}), with respect to embedding matrix `E', instead of input examples as given in equation (\ref{eq:2}) to calculate the amount of perturbation. We add this perturbation to the embedding matrix and the network goes through another forward pass. Finally, we calculate another classification loss against the adversarial example.  

\subsection{Contrastive Learning} 
Given a pair of clean and perturbed examples, we want to learn their representation similar to each other while learning different representations for the examples that are not from the same pair. To learn this representation, we leverage contrastive learning as a part of fine-tuning process. Similar to SimCLR\cite{chen2020simple}, we use in-batch negative examples and use its loss function as an objective function as well as project the final hidden state $h_[CLS]$ of both clean and adversarial examples to lower dimensions, i.e., 300. 
     
For a batch size of N clean examples and their corresponding perturbed examples, the are 2(N-1) negative pairs for each positive pair, i.e., all the remaining examples are negative examples for a positive pair. Contrastive loss function can be given as follows:

\begin{equation}
        \mathcal{L}_{ctr} = -log(\frac{\exp(sim(z_i, z_j/ \tau ))}{\sum_{k=1}^{2N} \mathbb{1} [k \neq i]  \exp(sim(z_i, z_j/ \tau ))})
     \end{equation}

where $sim(z_i, z_j)$ is the cosine similarity between vectors $z_i$ and $z_j$ and $\tau$ represents temperature hyperparameter. \\

Similar to\cite{pan2021improved}, we take the weighted average of two classification losses (for clean and its adversarial example) and the contrastive loss as given below:

 \begin{equation}
        \mathcal{L} = \frac{(1- \lambda )}{2} (\mathcal{L}_{CE_{1}} + \mathcal{L}_{CE_{2}}) + \lambda \mathcal{L}_{ctr}
     \end{equation}
where $\lambda$ controls the weightage of losses.

\section{Experiments}\label{exp}
In this section, first, we discuss the used dataset for training and evaluating our method. Then we give the pre-processing and training details for the method.
\subsection{Dataset}
We perform experiments on an extended version of the PHM2017 dataset provided by\cite{biddle2020leveraging} and split the dataset into 65\%, 15\%, and 20\% for the train, validation, and test set, respectively. This dataset contains data related to 10 diseases Alzheimer's, cancer, cough, depression, fever, headache, heart attack, migraine, Parkinson's, and stroke. Dataset detail is given in the Table \ref{tbl:stats1}.
\begin{table*}[!htbp]
\begin{center}
\caption{Extended version of PHM2017 dataset. There are a total of 15,742 total tweets in the dataset with 4,228 health-mention, 7,322 non-health mention, and 4,192 figurative mention tweets.}
\label{tbl:stats1}
\begin{tabular}{lllll}
\hline
Disease      & Tweet Count & Health Mention & Non-Health Mention & Figurative Mention \\
\hline \hline
Alzheimer’s  & 1,715                & 249                     & 1,374                       & 92                          \\ \hline
Cancer       & 1,691                & 302                     & 1,239                       & 150                         \\ \hline
Cough        & 1,452                & 331                     & 433                         & 688                         \\ \hline
Depression   & 1,579                & 517                     & 711                         & 351                         \\ \hline
Fever        & 1,484                & 517                     & 342                         & 625                         \\ \hline
Headache     & 1,429                & 791                     & 112                         & 526                         \\ \hline
Heart attack & 1,618                & 209                     & 349                         & 1,060                       \\ \hline
Migraine     & 1,519                & 904                     & 400                         & 215                         \\ \hline
Parkinson’s  & 1,568                & 153                     & 1,362                       & 53                          \\ \hline
Stroke       & 1,687                & 255                     & 1,000                       & 432                         \\
\hline \hline
Total        & 15,742               & 4,228                   & 7,322                       & 4,192    \\   
\hline
\end{tabular}

\end{center}

\end{table*}
\subsection{Pre-processing}
Each tweet goes through the pre-processing pipeline before going through the model. We first convert emojis in the tweet to text using  Python library. Then we remove all the user mentions, URLs, hashtags, and special characters. This preprocessing makes the emojis a part of the tweet text.
\subsection{Training Details}
We conduct experiments by using BERT\textsubscript{Large} and RoBERTa\textsubscript{Large} as baseline models. Then we apply contrastive adversarial training using these models. For all the experiments, we set a fixed learning rate of $1e^{-5}$, a maximum sequence length of 64, and fine models for 10 epochs. For BERT\textsubscript{Large} and RoBERTa\textsubscript{Large} as baselines, we search over a batch size of $\{16,24, 32\}$. For contrastive adversarial training, we perform grid search over $\lambda \in \{0.1,0.2,0.3,0.4,0.5\}$, $\epsilon \in \{0.02,0.005,0.001,0.0001\}$ , and $\tau \in \{0.05,0.06,0.07,0.08,0.09,0.1\}$. To compare the results with\cite{biddle2020leveraging} and\cite{khan2020improving}, we apply 10-fold cross validation, and choose the best validation hyper-parameters of batch size, $\lambda$, $\epsilon$, and $\tau$, and then report average validation results across 10-folds. 

\section{Results and Analysis}\label{raa}
We fine-tune two transformer models namely BERT\textsubscript{Large} and RoBERTa\textsubscript{Large} and use these models as the baseline for the task of HMC. For contrastive adversarial training, we use these two models with three losses, i.e., two classification losses (for cleaned and adversarial examples) and a contrastive loss, and take the weighted average of these losses.

Table~\ref{tbl:results1} shows the test set results on extended PHM2017 dataset. Contrastive adversarial method for BERT\textsubscript{Large} shows an improvement of $1.02\%$ over the baseline method, whereas for RoBERTa\textsubscript{Large} it shows an improvement of $0.61\%$ over the baseline method in terms of F1-score.  If we use only adversarial examples while training the model instead of contrastive adversarial, the classification performance decreases up to $0.35\%$  and $0.21\%$ for BERT\textsubscript{Large} and RoBERTa\textsubscript{Large}, respectively. Table~\ref{tbl:results2} shows the class-wise average F1-score for the test set. Adversarial only training improves the baseline score by $1.0\%$ for BERT\textsubscript{Large}, and $0.5\%$ for RoBERTa\textsubscript{Large} whereas contrastive adversarial training improves the F1 score by $1.5\%$, and $1.0\%$ over their respective baseline methods, respectively. These results suggest that contrastive learning helps the model learning better representation of the tweets.

In Table~\ref{tbl:results3}, we compare experimental results with other methods present in the literature. These results are directly comparable to only\cite{khan2020improving} because, at the time of download of the dataset, some tweets were not available, and hence dataset statistics only match to \cite{khan2020improving}. Results show that the contrastive adversarial method for BERT\textsubscript{Large} gains $4.15\%$ improvement of precision and $5.55\%$ improvement of recall over\cite{khan2020improving} method. It improves the F1 score by $5.05\%$ over the method in \cite{khan2020improving}. Contrastive adversarial method for RoBERTa\textsubscript{Large} improves precision, recall, and F1 score by $4.85\%$, $6.2\%$, and $5.8\%$ respectively over the method in \cite{khan2020improving}.

Deep learning models are black boxes in nature, i.e., it is unclear which features of the input influence the deep learning model to reach a decision. Hence, the use of deep learning in critical applications such as healthcare is questionable. European Union announced new regulations to implement a \enquote{right to explanation} which means a user can ask for the factors contributing to the decision of the deep learning model. Explainable AI\cite{gilpin2018explaining} focuses on providing the internals of the model in a human-understandable way to explain the factors influencing the model decision. Especially, various methods explain the model decision by feature, neuron, and layer importance, also known as layer attribution algorithms\cite{kokhlikyan2020captum}. In this paper, we visualize the important words that influence the model in reaching the classification decision using transformers-interpret library \cite{Pierse_Transformers_Interpret_2021}. In Table~\ref{tbl:analysis}, we plot some of the randomly selected tweets from the test set and analyze the importance of words in the classification decision of the model. First tweet \enquote{My wisdom teeth are coming in and they’re giving me the worse migraine ever weary face} is correctly classified by BERT\textsubscript{Large} contrastive, as health mention. The words \enquote{my, teeth, worse, and migraine} are influencing the model for classifying this tweet as health mention. The word \enquote{wisdom} is contributing towards non-health mention classification that makes sense because the word \enquote{wisdom} usually does not match the health mention scenarios. Overall score results in health mention classification by the model. The model BERT\textsubscript{Large} baseline, wrongly classifies this tweet as non-health mention. The words \enquote{are, giving, face} are resulting in the model's prediction of non-health mention because such words mostly appear in non-health mention tweets, and the model predicts based on these words. On the other hand, the words \enquote{teeth, ever, weary} are opposing the model prediction. BERT\textsubscript{Large} baseline, classifies the tweet \enquote{i rlly want a tattoo of a pretty crying woman smoking a cig but i don't want to manifest depression and addiction} as health mention whereas BERT\textsubscript{Large} contrastive, classifies it as non-health mention. The word \enquote{depression} influences BERT\textsubscript{Larger} base, to predict the tweet as health-mention whereas words such as \enquote{want, bit don't, and} influence it to predict as non-health mention. Word \enquote{want, pretty, crying, smoking, manifest} are attributing towards non-health mention classification by BERT\textsubscript{Large} contrastive, whereas, the word \enquote{depression} is opposing the classification as non-health mention. A health mention tweet \enquote{This is the first time in months I don’t have to work on a Saturday and I was SO excited to go out tn but ig god was like nah have a migraine instead sit in the dark on this Friday night.....bitch} is classified as health mention by RoBERTa\textsubscript{Large} contrastive, and as non-health mention by RoBERTa\textsubscript{Large} baseline. The words \enquote{migraine and god} mainly influence RoBERTa\textsubscript{Large} contrastive, for health mention classification. The reason for the contribution of \enquote{god} is the frequent usage of this word in health mention tweets in training data. In case of RoBERTa\textsubscript{Large} baseline, words \enquote{the, time, in, months, was, like, ah} contribute towards non-health mention classification whereas words \enquote{migraine, instead} oppose this prediction. The non-health mention tweet \enquote{I just straightened my hair out of depression wow look at me} is correctly classified by RoBERTa\textsubscript{Large} contrastive and misclassified by RoBERTa\textsubscript{Large} baseline. This is surprising that words \enquote{straightened, my, and hair} are suggesting the RoBERTa\textsubscript{Large} baseline model for health mention classification. Similarly, the contribution of words \enquote{wow and look} for health mention is surprising in the case of RoBERTa\textsubscript{Large} contrastive prediction because such words usually indicate figurative mention of the disease words.

\begin{figure*}
\centering
\caption{Embedding of baseline and contrastive adversarial training of both BERT\textsubscript{Large} and RoBERTa\textsubscript{Large} models for Health Mention Classification. These embeddings are plotted for the validation set of the extended PHM2017 dataset.}
\begin{tabular}{ccc}
\subfloat[BERT Baseline]{\includegraphics[width =3in]{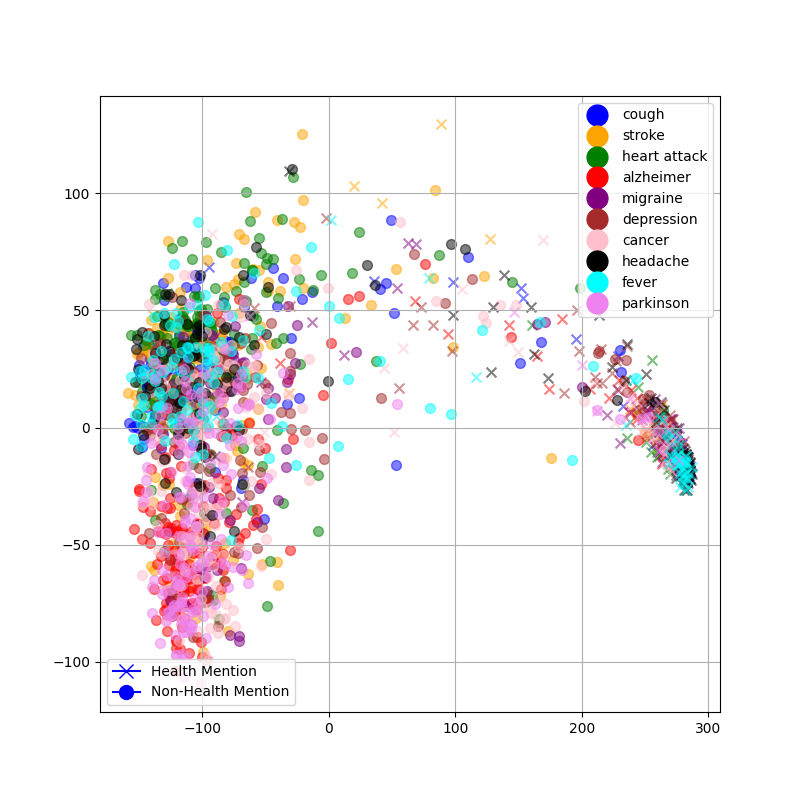}} &
\subfloat[RoBERTa Baseline]{\includegraphics[width = 3in]{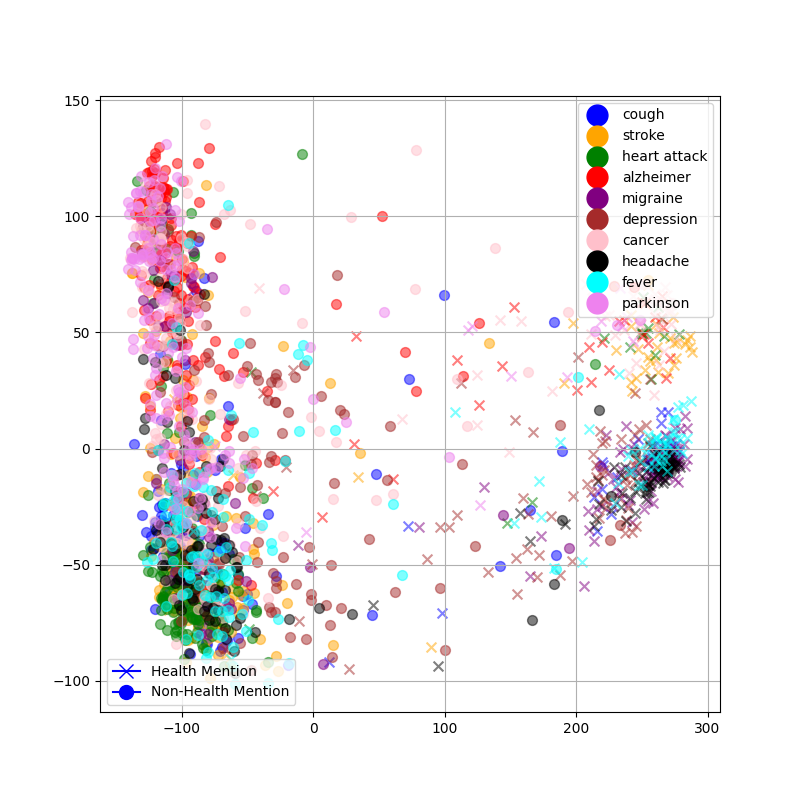}} \\\\
\subfloat[BERT Contrastive]{\includegraphics[width = 3in]{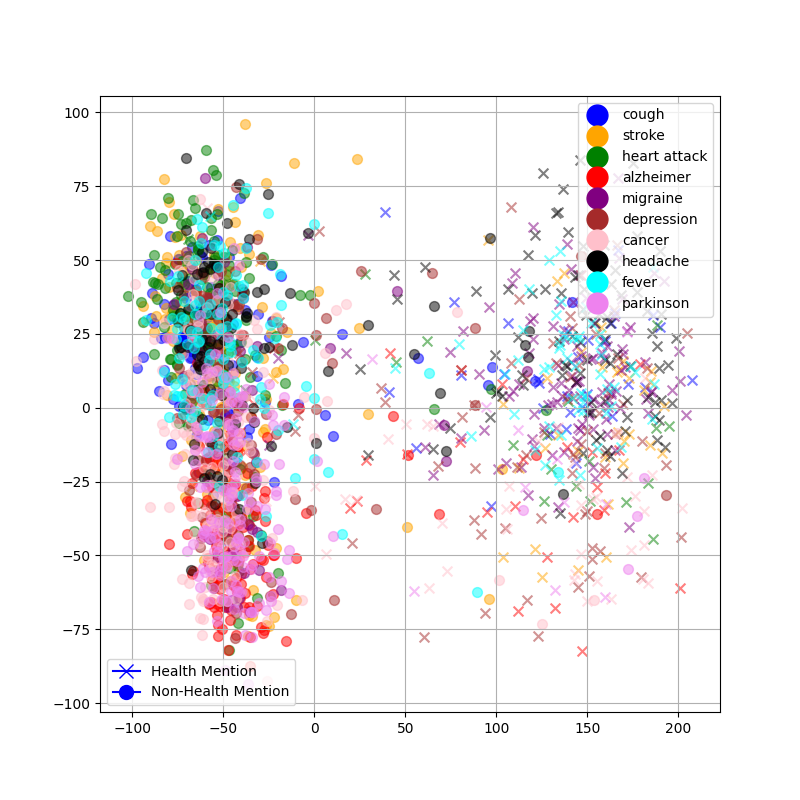}}& 
\subfloat[RoBERTa Contrastive]{\includegraphics[width = 3in]{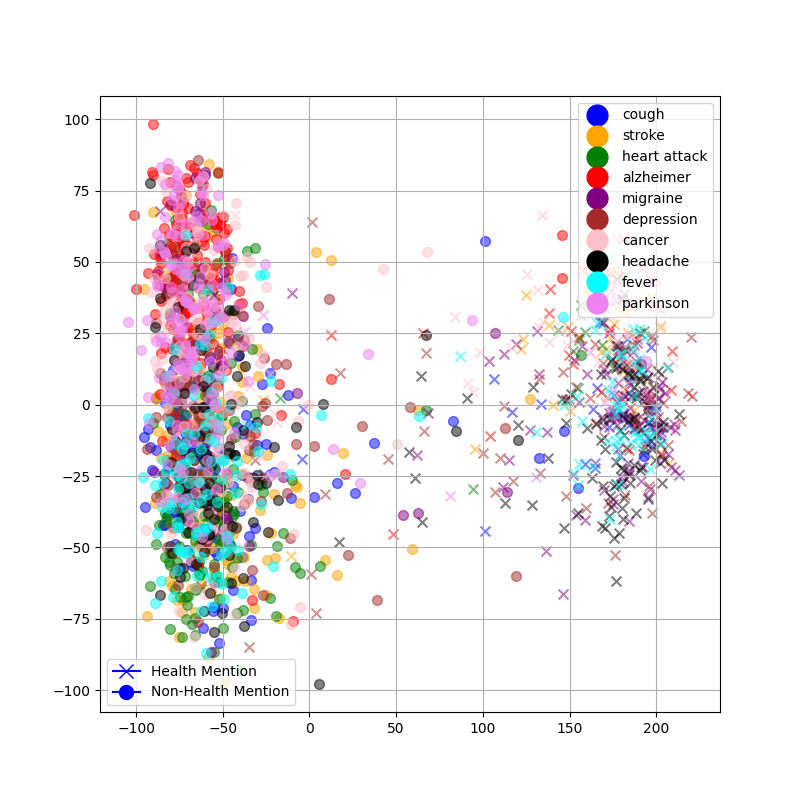}}
\end{tabular}

\label{fig-embedds}
\end{figure*}

These visualizations show that disease words themselves always contribute towards health mention prediction whereas other surrounding words support or oppose them for final prediction. In figure\ref{fig-embedds}, we plot the embedding of validation set learnt for the baseline and contrastive adversarial training of both BERT\textsubscript{Large} and RoBERTa\textsubscript{Large} models. The embeddings of non-health mention diseases for BERT\textsubscript{Large} contrastive are much more compact than BERT-baseline which helps the better performance of the model. Similarly, for RoBERTa\textsubscript{Large} baseline, we can see that many non-health mention examples lie in the health-mention embedding space which is not the case for RoBERTa\textsubscript{Large} contrastive embedding. Hence, RoBERTa\textsubscript{Large} contrastive performs better than RoBERTa\textsubscript{Large} baseline method.
 \begin{table}[!htbp]
\begin{center}
\caption{Results measured in terms of F1-score on the test set of extended version of PHM2017 dataset. Both BERT and RoBERTa using contrastive adversarial training improves the classification score over the baseline methods.}
\label{tbl:results1}
\begin{tabular}{llll}
\hline
Model     & Baseline & Adversarial & Contrastive Adversarial\\
\hline
BERT\textsubscript{Large} & 91.84 & 92.51 & \textbf{92.86}\\
\hline
RoBERTa\textsubscript{Large} & 93.13 & 93.53 &  \textbf{93.74}\\
\hline
\end{tabular}

\end{center}

\end{table}

 \begin{table}[!htbp]
\begin{center}
\caption{Results showing class-wise average F1-score on the test set of an extended version of PHM2017 dataset for baseline, adversarial, and contrastive adversarial training.}
\label{tbl:results2}
\begin{tabular}{llll}
\hline
Model     & Baseline & Adversarial & Contrastive Adversarial\\
\hline
BERT\textsubscript{Large} & 91.5 & 92.5 & \textbf{93}\\
\hline
RoBERTa\textsubscript{Large} & 93 & 93.5 &  \textbf{94}\\
\hline
\end{tabular}

\end{center}

\end{table}

 \begin{table}[!htbp]
\begin{center}
\caption{Results showing class-wise average Precision, Recall, and F1-score  on the 10-fold validation of an extended version of PHM2017 dataset and comparing them to the existing methods. Our method is directly comparable to only Khan et al. method\cite{khan2020improving} due to the same distribution of the dataset.}
\label{tbl:results3}
\begin{tabular}{llll}
\hline
Method     & Precision & Recall & F1 Score\\
\hline
Jiang et al.\cite{jiang2018identifying} & 72.1 & 95 & 81.8\\
\hline
Karisani et al.\cite{karisani2018did}  & 75.2 & 89.6 & 81.8\\
\hline
Biddle et al.\cite{biddle2020leveraging}  & 75.6 & 92 & 82.9\\
\hline
Khan et al.\cite{khan2020improving}  & 89.1 & 88.2 & 88.4\\
\hline

Contrastive Adversarial on BERT\textsubscript{Large}  & 93.25 & 93.75 & 93.45\\
\hline

Contrastive Adversarial on RoBERTa\textsubscript{Large} & \textbf{93.95} & \textbf{94.4} & \textbf{94.2}\\
\hline
\end{tabular}
\end{center}

\end{table}

 \begin{table*}[!htbp]
\begin{center}
\caption{Visualizations showing important words that influence by the model for its classification decision. Green highlighted words are those which contributed to the model classification decision. Red highlighted words are those which opposed the model decision.}
\begin{tabular}{llll}
\hline
True Label & Prediction &Model     & Word Importance\\
\hline
\\[-1em]
HM & NHM & BERT\textsubscript{Large} Baseline  & \includegraphics[width=10cm, height=0.35cm]{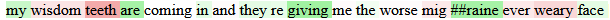} \\

 & HM & BERT\textsubscript{Large} Contrastive & \includegraphics[width=10cm, height=0.35cm]{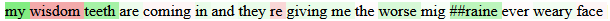} \\
\hline
\\[-1em]
NHM & HM & BERT\textsubscript{Large} Baseline & \includegraphics[width=10cm, height=0.35cm]{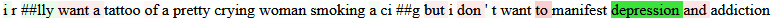} \\

 & NHM & BERT\textsubscript{Large} Contrastive & \includegraphics[width=10cm, height=0.35cm]{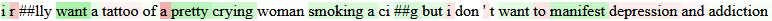} \\
\hline
\\[-1em]
HM & NHM & RoBERTa\textsubscript{Large} Baseline & \includegraphics[width=10cm, height=0.9cm]{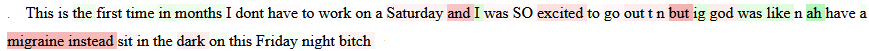} \\

 & HM & RoBERTa\textsubscript{Large} Contrastive & \includegraphics[width=10cm, height=0.9cm]{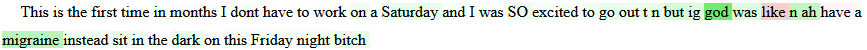} \\
\hline
\\[-1em]
NHM & HM & RoBERTa\textsubscript{Large} Baseline & \includegraphics[width=10cm, height=0.25cm]{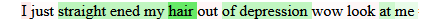} \\

 & NHM & RoBERTa\textsubscript{Large} Contrastive & \includegraphics[width=10cm, height=0.25cm]{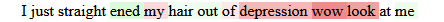} \\
\hline
\end{tabular}

\label{tbl:analysis}

\end{center}

\end{table*}

\section{\uppercase{Conclusion}}\label{conc}
\label{sec:conclusion}

In this paper, we utilize contrastive adversarial training for the health mentioning classification task that acts as a regularizer. We evaluate the method performance on an extended version of the PHM2017 dataset and compare the results with baseline and SOTA methods. Experiments results show that adversarial and contrastive training significantly improves the tweet health mentioning classification performance over the baseline methods. We visualize some of the tweets that are correctly classified by the contrastive adversarial training and misclassified by baseline models to understand the classification decisions made by these models.

\EOD
\bibliographystyle{unsrt}
\bibliography{refrences}
\end{document}